\documentclass[conference]{IEEEtran}
\IEEEoverridecommandlockouts
\usepackage{cite}
\usepackage{amsmath,amssymb,amsfonts}
\usepackage[ruled,linesnumbered]{algorithm2e}
\usepackage{graphicx}
\usepackage{textcomp}
\usepackage{xcolor}
\usepackage{stfloats}
\usepackage{array}
\usepackage[caption=false,font=normalsize,labelfont=sf,textfont=sf]{subfig}
\usepackage{verbatim}
\usepackage[a-2b]{pdfx}\usepackage[a-2b]{pdfx}
\usepackage{multirow}
\usepackage{longtable}
\usepackage{amsthm}

\def\BibTeX{{\rm B\kern-.05em{\sc i\kern-.025em b}\kern-.08em
    T\kern-.1667em\lower.7ex\hbox{E}\kern-.125emX}}
\begin{document}

\title{Dual-channel Heterophilic Message Passing for Graph Fraud Detection}

\author{

\IEEEauthorblockN{
    Wenxin Zhang \IEEEauthorrefmark{1}\textsuperscript{1},
    Jingxing Zhong \IEEEauthorrefmark{2}\textsuperscript{1}, 
    Guangzhen Yao \IEEEauthorrefmark{3}\textsuperscript{1},
    Renda Han \IEEEauthorrefmark{4}\textsuperscript{1}, 
    Xiaojian Lin \IEEEauthorrefmark{5}\textsuperscript{1},
    Lei Jiang \IEEEauthorrefmark{1}\textsuperscript{2}, \\
    Zeyu Zhang \IEEEauthorrefmark{6}\textsuperscript{1}
    and Cuicui Luo\IEEEauthorrefmark{1}\textsuperscript{3}\IEEEauthorrefmark{7}
}

\IEEEauthorblockA{
    \IEEEauthorrefmark{1}University of Chinese Academy of Science, Beijing, China \\
    \textsuperscript{1}zwxzhang12@163.com, \textsuperscript{2}jianglei\_0123@163.com, \textsuperscript{3}luocuicui@ucas.ac.cn
}

\IEEEauthorblockA{
    \IEEEauthorrefmark{2}Fuzhou University, Fuzhou, China
    \textsuperscript{1}832303225@fzu.edu.cn
}

\IEEEauthorblockA{
    \IEEEauthorrefmark{3}Northeast Normal University, Changchun, China
    \textsuperscript{1}yaoguangchen@nenu.edu.cn
}

\IEEEauthorblockA{
    \IEEEauthorrefmark{4} Hainan University, Haikou, China \textsuperscript{1}hanrenda@hainanu.edu.cn
}

\IEEEauthorblockA{
    \IEEEauthorrefmark{5}Tsinghua University, Beijing, China
    \textsuperscript{1}chrslim@connection.hku.hk
}

\IEEEauthorblockA{
    \IEEEauthorrefmark{6}The Australian National University, Canberra, Australia \textsuperscript{1}steve.zeyu.zhang@outlook.com
}

\IEEEauthorblockA{
    \IEEEauthorrefmark{7}Corresponding Author
}
}

\maketitle

\begin{abstract}
Fraudulent activities have significantly increased across various domains, such as e-commerce, online review platforms, and social networks, making fraud detection a critical task. Spatial Graph Neural Networks (GNNs) have been successfully applied to fraud detection tasks due to their strong inductive learning capabilities. However, existing spatial GNN-based methods often enhance the graph structure by excluding heterophilic neighbors during message passing to align with the homophilic bias of GNNs. Unfortunately, this approach can disrupt the original graph topology and increase uncertainty in predictions. To address these limitations, this paper proposes a novel framework, Dual-channel Heterophilic Message Passing (DHMP), for fraud detection. DHMP leverages a heterophily separation module to divide the graph into homophilic and heterophilic subgraphs, mitigating the low-pass inductive bias of traditional GNNs.  It then applies shared weights to capture signals at different frequencies independently and incorporates a customized sampling strategy for training. This allows nodes to adaptively balance the contributions of various signals based on their labels. Extensive experiments on three real-world datasets demonstrate that DHMP outperforms existing methods, highlighting the importance of separating signals with different frequencies for improved fraud detection. The code is available at \url{https://github.com/shaieesss/DHMP}.
\end{abstract}


\section{Introduction}
The rapid evolution of Internet services has undeniably enriched our daily lives, yet it has concurrently introduced a spectrum of fraudulent activities. In reality, fraudsters often disguise themselves as legitimate users during malicious undertakings, effectively burying their illicit activities within vast data flows. These deceptive practices present significant risks to internet security \cite{DengWLWC23}, financial security \cite{SplitGNN}, and review management systems \cite{H2-FDetector}. However, online interactions, such as retweets, reviews, and transactions, can naturally be mapped onto complex graph-structured data by representing data objects and their interactions as node entities and connections, which makes these actions invariably traceable. To leverage the powerful representational capabilities of complex graph structures, both industrial practitioners and researchers have increasingly turned to spatial Graph Neural Networks (GNNs) for fraud detection.

\begin{figure}
    \centering
    \includegraphics[width=0.9\linewidth]{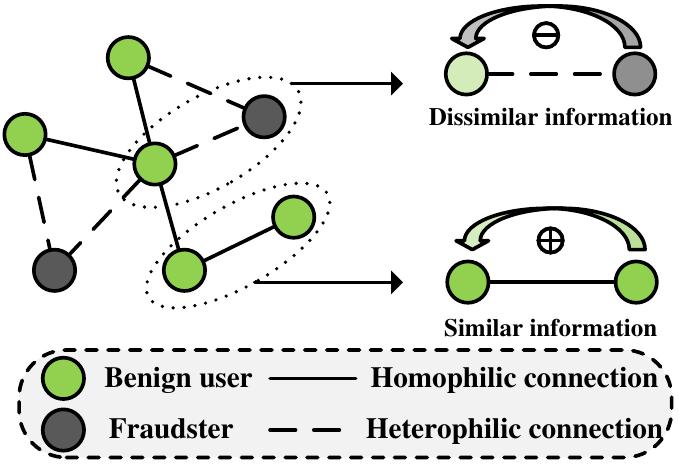}
\caption{Illustration of two key challenges in fraud detection: (1) The pronounced disparity between the number of fraudulent and benign users, with fraudsters often being significantly outnumbered, increases the difficulty of identifying and mitigating fraudulent activities. (2) Homophilic connections promote the blending of information between entities within the same categories, while heterophilic connections blur the boundaries between different classes, further complicating the detection of deceitful interactions.}
    \label{intro}
\end{figure}

Deploying spatial GNNs for fraud detection faces two major challenges: label imbalance and the intricate coexistence of heterophily and homophily. On the one hand, network attackers often employ advanced strategies to mimic legitimate behaviors, subtly embedding a small number of fraudulent nodes within the broader context of a target graph. This deliberate embedding conceals their malicious activities. For instance, as illustrated in Fig. \ref{intro}, two fraudsters disproportionately hide themselves among five benign users. Unfortunately, this imbalance makes it more difficult to extract distinguishing features of fraudulent activity within the network. Addressing this challenge requires implementing targeted strategies to mitigate the effects of label imbalance, which introduces additional complexity in model design.

On the other hand, the interaction between heterophily and homophily poses a significant challenge to the homophily-based inductive bias inherent in conventional Graph Neural Networks (GNNs). Traditional GNNs propagate and aggregate nodes' features within their neighborhoods through various mechanisms, such as summation and averaging. These operations effectively act as low-pass filters, smoothing signals to emphasize similarities among connected nodes. However, fraudsters often attempt to camouflage their identities by strategically forming extensive connections with benign users, aiming to blend in and appear normal. As illustrated in Fig. \ref{intro}, fraudulent users strategically establish connections with many benign users, generating substantial volumes of high-frequency and heterophilic data. This interaction pattern reduces the clarity of feature distinctions, making it harder to differentiate fraudulent nodes. Consequently, fraudulent nodes, specifically positioned within benign communities, exhibit heterophily, while the surrounding context of benign nodes displays homophily. Addressing these challenges requires developing GNN models that effectively handle this complex interplay, which is essential for improving fraud detection.

Numerous spatial-based models incorporate sophisticated sampling techniques to mitigate the adverse effects of heterophilic neighbors during the aggregation phase \cite{CARE-GNN, DiG-In-GNN}. By employing pruning strategies, these models ensure that anchor nodes exclusively aggregate features from homogeneous neighbors, thereby significantly enhancing the proximity of node features within the same category. These approaches effectively address the challenges posed by data imbalance and heterophily within graph data. However, they introduce hard manipulations into the message propagation process, thereby disrupting the integrity of the endogenous topology of the original graph. Additionally, the reliance on similarity predictions to determine edge weights or graph pruning probabilities is prone to cumulative errors, which can propagate and amplify the inaccuracies and ultimately lead to significant deviations from the desired and expected outcomes.

Based on the above analysis, we propose a novel Dual-channel Heterophilic Message Passing (DHMP) framework for fraud detection problems. Specifically, DHMP first utilizes a heterophily separation module to discern homophilic and heterophilic connections within the graph. Subsequently, DHMP proceeds to aggregate homophilic and heterophilic information by dual independent channels with correlated parameters, obtaining the complementary information embedded within these signals with different frequencies. Following this aggregation phase, DHMP assembles the representations derived from different frequencies and multifaceted relations, generating cohesive and comprehensive embeddings. Ultimately, DHMP predicts the labels of nodes through a normalized classifier.

For the heterophily challenge, DHMP introduces a heterophily separation module to split the original graph into homophilic and heterophilic subgraphs and independently proceeds with message aggregation. This way, DHMP can separately capture assortative behavior and camouflage information, conquering the low-pass filtering of traditional spatial GNNs.
For the data imbalance challenge, by disentangling heterophilic edges, nodes are empowered to apportion the influence of homogeneous and heterogeneous nodes dynamically during the process of message propagation. This mechanism facilitates an adaptive learning paradigm for nodes, enabling them to acquire features that are intrinsically congruent with their respective labels. Consequently, this approach mitigates, to a significant degree, the challenges associated with feature learning that arise from label insensitivity, thereby enhancing the robustness and accuracy of the learning process. In addition, DHMP adopts a tailored sampling strategy to ensure the convergence of the training process. In summary, the major contributions are as follows:
\begin{itemize}
\item  A novel Dual-channel Heterophilic Message Passing (DHMP) framework is proposed to address the data imbalance and heterophily problems in fraud detection.
\item  A heterophily separation module is proposed to perceive high-frequency signals and decouple heterophily in the graph.
\item A re-scale residual message propagation mechanism is proposed to capture graph signals in different frequencies, which adaptively preserves original characteristics through a weighted residual gate and re-scales the features via the node degree.
\item Extensive experiments are conducted on three public datasets, verifying the advancement of DHMP compared to the state-of-the-art baselines.
\end{itemize}

\textbf{Related} \textbf{Work} GNNs-based fraud detection methodologies can exploit graphs' intrinsic topological characteristics, leveraging information dissemination among the nodes to capture critical characteristics of fraudsters \cite{CARE-GNN}. For example, IMINF \cite{IMINF} utilizes a multi-scale neighbor sampling mechanism to disseminate pertinent information. DIG-In-GNN \cite{DiG-In-GNN} integrates guidance information to refine the message-passing procedure, enhancing the fidelity of information exchange. ASA-GNN \cite{ASA-GNN} implements adaptive sampling techniques to sift out noisy nodes and propagate more informative and representative data. However, they often yield biased feature representations by disrupting structural information and neglecting heterophilic information, leading to a deterioration in the model's recognition performance. Ubiquitous heterophilic information can provide crucial insights into fraudulent interaction characteristics and behavioral patterns.

\section{PRELIMINARIES}
This section provides imperative definitions and detailed descriptions of the research problem.
\subsection{Definitions}
\textbf{Definition} $\mathbf{1}$: $(Graph)$: Let $\mathcal{G} = (\mathcal{V}, \mathcal{X}, \mathcal{E}, \mathcal{Y})$ denotes a directed graph. $\mathcal{V} = \{v_{1}, v_{2}, ...,  v_{N}\}$ denotes the node set and $|\mathcal{V}|=N$. $\mathcal{X} \in \mathbb{R}^{N \times d}$ is the corresponding feature matrix of node set $\mathcal{V}$, where $d$ is the original representation dimension. $\mathcal{E}$ is the edge set and $e_{uv} \in \mathcal{E}$ denotes an edge from node $u$ to node $v$. $\mathcal{Y}=\{0, 1\} \in \mathbb{R}^{N \times 1}$ represents the label set. $y_v \in \mathcal{Y} = 0$ indicates that the node $v$ is a benign entity, otherwise the node $v$ is an anomaly.

\textbf{Definition} $\mathbf{2}$: $(Multi-relation graph)$: A multi-relation graph can be defined as $\mathcal{G} = (\mathcal{V}, \mathcal{X}, \mathcal{E}_{r} |^{R}_{r=1}, \mathcal{Y})$ where $R$ is the number of relations.
\subsection{Problem statement}
Considering a multi-relation graph $\mathcal{G}$, the fraud detection problem is geared to identify fraudulent entities in the graph $\mathcal{G}$ by elaborating a model $f_{\theta}(\cdot)$ with adaptive parameters $\theta$. The model $f_{\theta}(\cdot)$ captures discriminative latent feature space $\Omega$ for nodes through specific transformation and optimization according to node features $\mathcal{X}$, their connections $\mathcal{E}$ and labels $\mathcal{Y}$.

\section{METHODOLOGY}
In this section, we present our fraud detection framework, DHMP, and then elaborate on its components and training process.
\subsection{framework}
\begin{figure*}
    \centering
    \includegraphics[width=0.95\linewidth]{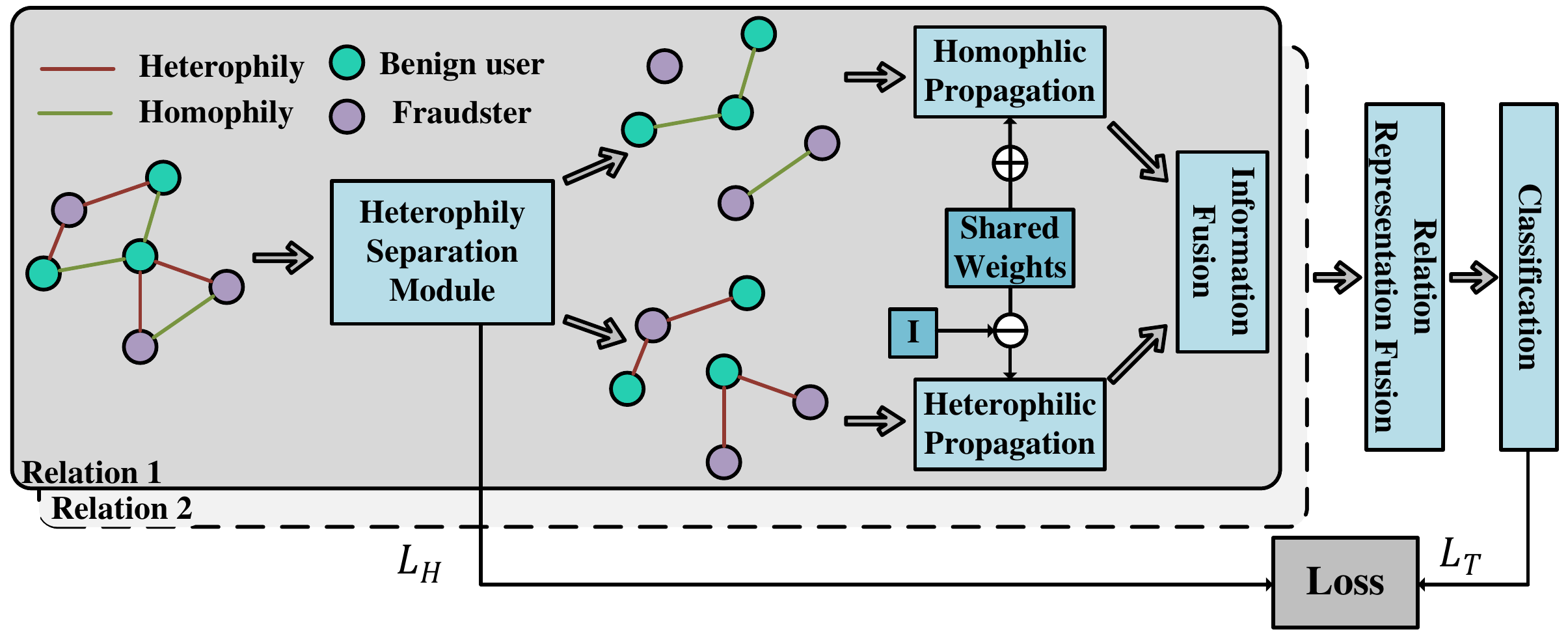}
    \caption{The framework of DHMP. First, DHMP leverages a heterophily separation module to split the low-pass and high-pass signals in the original graph. Then, DHMP leverages the re-scaled residual message aggregation module to propagate the homophilic and heterophilic information. Finally, DHMP fuses the relational information and detects the fraudulent entities. DHMP is trained using the joint loss functions, including classification and heterophily loss.}
    \label{framework}
\end{figure*}

The framework of DHMP is shown in Fig. \ref{framework}. First, DHMP leverages a heterophily separation module to perceive and split homophilic and heterophilic connections. Then, DHMP respectively executes homophilic and heterophilic message aggregation to generate unique information through a re-scaled residual message-passing mechanism and integrates information about homophily and heterophily. Next, DHMP assembles the representations from diverse relations. Last, DHMP normalizes the embeddings and predicts fraudsters through a classifier.
\subsection{Heterophily Separation}
The aggregation of neighborhood features in traditional GNNs is facilitated by applying a low-pass filter, a methodology primarily suited for graphs exhibiting high homogeneity. However, when extended to graphs with high heterophily, these models often inadvertently smooth out the distinctions among features belonging to different categories, ultimately leading to performance decline. We introduce a novel heterophily separator to address this limitation and effectively discern heterophily within graph topologies. The primary objective of the separator is to perceive the heterophilic and homophilic connections within the graph precisely by utilizing the labels of source and target nodes in the training dataset. 

Specifically, the heterophily separator is formulated as a binary classification model. This model harnesses the features of both the source node $u$ and the target node $v$ to predict the type of the connection $e_{uv}$. The implementation of this separator is grounded in the Multi-Layer Perceptron (MLP) architecture, empowering the model to differentiate between various edge types within the graph with high efficacy. For an edge $e_{uv}$, we first leverage feature projection for the corresponding nodes:
\begin{equation}
\label{edge0}
   h = \sigma(W_s \cdot x + b_s),
\end{equation}
where $\sigma(\cdot)$ is an activation function, $W_s$ and $b_s$ are learnable weights of the linear transformation, and $x$ denotes the original features of node. Then we calculate the sign of the edge by leveraging $tanh(\cdot)$ non-linear activation function:
\begin{equation}
\label{edge2}
   \kappa_{uv} = tanh(W_h[h_u||h_v||(h_u - h_v)]),
\end{equation}
where $h_u$ and $h_v$ indicate the representations of the neighboring node and anchor node, $W_h$ are learnable parameters, and $[\cdot||\cdot]$ represents the concatenation function.

To partition the original graph into a homophilic subgraph $\mathcal{G}_{p}$ and a heterophilic subgraph $\mathcal{G}_{n}$, we employ the predictive outcomes associated with all edges within the graph. Specifically, the homophilic subgraph exclusively comprises connections anticipated to bespeak homophily, whereas the heterophilic subgraph solely contains connections predicted to unfold heterophily.

Accurate awareness of heterophily holds predominant significance for following procedures, as it directly impacts the quality of capturing signals with different frequencies. To address this imperative, we formulate a specialized auxiliary loss function based on the training edge set $\mathcal{E}_{tr}$:
\begin{equation}
\label{edge3}
   \mathcal{L}_{H} = \frac{1}{|\mathcal{E}_{tr}|} \sum_{e_{uv} \in \mathcal{E}_{tr}} max(1 - \kappa_{uv} \cdot y_{e_{uv}}, 0),
\end{equation}
\begin{equation}
\label{edge4}
y_{e_{uv}} = \left\{
\begin{aligned}
     -1 &\quad \mbox{if} & \quad y_u=y_v\\
      1 &\quad \mbox{if} & \quad y_u\neq y_v,
\end{aligned}
\right.
\end{equation}
where $y_{e_{uv}}$ is the label of edge $e_{uv}$, and it can be obtained according to the labels of the nodes on both sides.

\subsection{Re-scaled Residual Message Aggregation}
To capture different signals in the graph, we naturally leverage a graph transformation filter $W_f$ for the homophilic connection and design $I - W_f$ as the heterophilic filter. In this way, the transformed features will contain signals in different frequencies.

\subsubsection{Homophilic Propagation}
Formally, given an anchor node $u$ and its homophilic neighboring node set $\mathcal{V}_{p}(u)$ in $\mathcal{G}_{p}$, we first leverage the graph filter $W_f$ to transform the neighboring nodes of anchor node $u$:
\begin{equation}
\label{ma1p}
   \hat{h}_v^{+} = W_f \cdot h_v + b_{p1},
\end{equation}
where $h_v$ denotes the features of neighboring node $v \in \mathcal{V}_{p}$ and $b_{p1}$ is learnable parameters. 

To preserve the original characteristics of neighboring nodes, we design a weighted residual network to generate integrated neighboring information:
\begin{equation}
\label{ma2p}
   \Bar{h}_v^{+} = \phi(W_p(\epsilon \cdot h_v + \hat{h}_v^{+})+ b_{p2}),
\end{equation}
where $W_p$ and $b_{p2}$ are learnable parameters, $\phi(\cdot)$ denote LeakyReLU activation function and $\epsilon$ is a balance hyperparameter.

To enhance the node representation learning, we implement re-scaled neighboring information aggregation based on the embeddings of all neighbouring the set $\mathcal{V}_{p}$ and the representations of the anchor node $h_u$ to obtain the updated representations of the anchor node:

\begin{equation}
\label{ma3p}
   z_u^{+} = h_u + \sum_{v \in \mathcal{V}_{p}(u)}\frac{\Bar{h}_v^{+}}{\sqrt{1 + d_u^{+} \cdot d_v^{+}}},
\end{equation}
where $d_u^{+}$ and $d_v^{+}$ are respectively the degree of anchor node $u$ and its neighbor $v$ in subgraph $\mathcal{G}_{p}$, and $h_u$ is the representation of anchor node. 

We explain the incentive of the re-scale factors. For anchor nodes, the larger the degree of the node, the more information sources it has, and the smaller the proportion of each piece of information. For neighboring nodes, the greater the degree of the node, the less information it propagates to a single branch.

\subsubsection{Heterophilic Propagation}
For heterophilic subgraph $\mathcal{G}_{n}$, we define anchor node $u$ and its heterophilic neighboring node set $\mathcal{V}_{n}(u)$. Similar to Eq. \ref{ma1p} and Eq. \ref{ma2p}, we first derive the message from neighbors according to the following formulations:
\begin{equation}
\label{ma1n}
   \hat{h}_v^{-} = \sigma((I - W_f) \cdot h_v + b_{n1}),
\end{equation}
\begin{equation}
\label{ma2n}
   \Bar{h}_v^{-} = \phi(W_n(\epsilon \cdot h_v + \hat{h}_v^{-})+ b_{n2}),
\end{equation}
where $b_f$, $W_n$ and $b_{n2}$ are learnable parameters, $h_v$ represents the representations of neighboring node $v \in \mathcal{V}_{n}(u)$.

Then, we propagate the messages from neighbors and update the features of center node $u$:
\begin{equation}
\label{ma3n}
   z_u^{-} = h_u + \sum_{v \in \mathcal{V}_{n}(u)}\frac{\Bar{h}_v^{-}}{\sqrt{1 + d_u^{-} \cdot d_v^{-}}},
\end{equation}
where $d_u^{-}$ and $d_v^{-}$ are respectively the degree of anchor node $u$ and its neighbor $v$ in subgraph $\mathcal{G}_{n}$.
\subsection{Frequency Information Fusion}
Given the updated homophilic and heterophilic representations, treating node representations independently under specific connections may overlook their interaction, leading to biased information and performance deterioration. A direct way is to concatenate the representations across different subgraphs, which can be implemented as follows:
\begin{equation}
\label{fusion1}
   z_u = W_i[z_u^{+}||z_u^{-}] + b_i,
\end{equation}
where $ [\cdot||\cdot] $ denotes the concatenation function, and $W_i$ and $b_i$ are learnable fusion parameters. However, the vanilla concatenation fails to discriminate the deviation between the embeddings. Hence, we modify the Eq. \ref{fusion1} to enhance the ability to capture latent dependencies between the embeddings, which can be shown as follows:
\begin{equation}
\label{fusion2}
   z_u = Norm(\phi(W_i[z_u^{+}||z_u^{-}||(z_u^{+} - z_u^{-})] + b_i)),
\end{equation}
where $Norm(\cdot)$ denotes layer normalization to boost the expressive ability of node features.

\subsection{Relation Representation Fusion}
In practical applications, fraud graphs typically encompass many diverse relationships. Once representations for each of these relationships have been obtained, the subsequent step involves integrating the node representations derived from these various relationships to construct a definitive embedding for the nodes. For conciseness, we define the output embeddings of the subgraph information fusion module under relation $r$ as $z_u^r$. We aggregate the node features from different relations to acquire the final embeddings of node $u$:
\begin{equation}
\label{fusion3}
   \Tilde{z}_u = ||_{r=1}^{R}(z_u^{r}),
\end{equation}
where $||(\cdot)$ is concatenation function.

\subsection{Classifier and Training}
We leverage the cross-entropy loss function to train the proposed model. Given the training set $\mathcal{V}_{tr}$, the classification loss function is defined as follows:
\begin{equation}
\label{cross loss}
  \mathcal{L}_{T}= -\sum_{u \in \mathcal{V}_{tr}}[(1 - y_u)log(1 - m_u) + y_vlog(m_u)],
\end{equation}
\begin{equation}
\label{cross loss1}
  m_u= softmax(\Tilde{z}_u),
\end{equation}
where $y_u$ is the label of node $u$.

The overall loss function combines the classification loss function and the auxiliary loss function:
\begin{equation}
\label{overall loss}
  \mathcal{L} = \mathcal{L}_{T} + \lambda\mathcal{L}_{H},
\end{equation}
where $\lambda$ is hyperparameter.

We incorporate a tailored sampling strategy during the training phase to further moderate the pervasive data imbalance challenge in fraud detection. When computing the node classification loss, this strategy involves selecting an equitable number of benign nodes as fraudulent nodes. Analogously, to calculate the edge classification loss, we ensure that homophilic edges are represented proportionately to heterophilic edges. Algorithm \ref{training} illustrates all the steps of DHMP.

\begin{algorithm}[!htbp]
\caption{The training process of DHMP}
\label{training}
\LinesNumbered 
\KwIn{A multi-relational graph $\mathcal{G} = (\mathcal{V}, \mathcal{X}, \mathcal{E}_{r} |^{R}_{r=1}, \mathcal{Y})$ for training; Learning rate $lr$; Training epochs $N_{epoch}$\; 
\For{epoch $\leftarrow$ 1, $\ldots$, $N_{epoch}$}{
    \For{r $\leftarrow$ 1, $\ldots$, $R$}{
        Generate homophilic subgraph $\mathcal{G}^{+}$ and heterophilic subgraph $\mathcal{G}^{-}$ under relation $r$\;
        Compute heterophilic separation loss $\mathcal{L}_{H}$ $\leftarrow$ Eq. \ref{edge3}\;
        \For{node $u \in \mathcal{V}$}{
            Obtain its homophilic neighbors $\mathcal{N}_{p}(u)$ in homophilic subgraph $\mathcal{G}^{+}$ and heterophilic neighbors $\mathcal{N}_{n}(u)$ in heterophilic subgraph $\mathcal{G}^{-}$\;
            \For{node $v \in \mathcal{N}_{p}(u)$}{
                Execute homophilic propagation and obtain embeddings $z_u^{-}$ $\leftarrow$ Eq. \ref{ma1p} - Eq. \ref{ma3p}\;
                
            }
            \For{node $v \in \mathcal{N}_{n}(u)$}{
                Execute heterophilic propagation and obtain embeddings $z_u^{-}$ $\leftarrow$ Eq. \ref{ma1n} - Eq. \ref{ma3n}\;
            }
            Obtain $z_u$ under single relation $\leftarrow$ Eq. \ref{fusion2}\;
        }
    }
    Obtain final multi-relational embeddings $\Tilde{z}_u$ $\leftarrow$ Eq. \ref{fusion3}\;
    Calculate classification loss $\mathcal{L}_{T}$ $\leftarrow$ Eq. \ref{cross loss}\;
    Calculate overall loss $\mathcal{L}$ $\leftarrow$ Eq. \ref{overall loss}\;
    Update model parameters\;	
}}
\KwOut{The final representations of nodes}
\end{algorithm}

\section{EXPERIMENTS}
In this section, we evaluate the performance of DHMP on three real-world datasets to answer the following research questions:
\begin{itemize}
\item \textbf{RQ1}: How does the performance of DHMP compare to that of the advanced methods in fraud detection problems?
\item \textbf{RQ2}: How do the individual modules of DHMP contribute to its overall performance?
\item \textbf{RQ3}: What is the influence of differing model parameters on the operational effectiveness of DHMP?
\item \textbf{RQ4}: What is the influence of differing model parameters on the operational effectiveness of DHMP?
\end{itemize}
\subsection{Experimental setup}
\subsubsection{Datasets}
We conduct comparative experiments on three public fraud detection datasets to verify the performance of DHMP. The statistical descriptions of datasets are listed in Table \ref{DATASETS}.
\begin{itemize}
\item YelpChi: \cite{CARE-GNN} This dataset describes fraudulent review comments on the Yelp platform, specifically those intended to promote or demote particular products or businesses. The dataset has three relations: R-U-R, R-S-R, and R-T-R.
\item Amazon: \cite{H2-FDetector} This dataset is designed to detect users compensated for generating counterfeit reviews for musical instruments on Amazon.com. The dataset has three relations: U-P-U, U-S-U, and U-V-U.
\item FDCompCN: \cite{SplitGNN} This dataset identifies financial statement fraud of Chinese companies from the China Stock Market and Accounting Research database. The dataset has three relations: C-I-C, C-S-C, and C-P-C.
\end{itemize}

\begin{table}[!htbp]
	\centering
	\caption{Descriptions of datasets}
	\label{DATASETS}
    \resizebox{0.45\textwidth}{!}{
	\begin{tabular}{c c c c c c} 
		\hline
		 Dataset & \#Node & Fraud(\%) & Dimension & Relation & \#Edge\\\hline
		 \multirow{3}{*}{YelpChi} & \multirow{3}{*}{45954} & \multirow{3}{*}{14.53\%}  &\multirow{3}{*}{32} & R-U-R & 98630\\
             ~ & ~ & ~  &~ & R-T-R & 1147232\\
             ~ & ~ & ~  &~ & R-S-R & 7693958\\
		 \multirow{3}{*}{Amazon} & \multirow{3}{*}{11944} & \multirow{3}{*}{6.87\%}  &\multirow{3}{*}{24} & U-P-U & 351216\\
             ~ & ~ & ~  &~ & U-S-U & 7132958\\
             ~ & ~ & ~  &~ & U-V-U & 2073474\\
		 \multirow{3}{*}{FDCompCN} & \multirow{3}{*}{5317} & \multirow{3}{*}{10.5\%}  &\multirow{3}{*}{57} & C-I-C & 5686\\
             ~ & ~ & ~  &~ & C-P-C & 760\\
             ~ & ~ & ~  &~ & C-S-C & 1043\\\hline
	\end{tabular}
    }
\end{table}

\subsubsection{Baselines}
We compare the DHMP with thirteen baselines to demonstrate the superiority of our model, which contains two shallow methods, four GNNs, and seven GNN-based fraud detection approaches.
\begin{itemize}
\item MLP: MLP is a classical neural network basic architecture. 
\item XGBoost: \cite{XGBoost}: XGBoost is an efficient and flexible tree structure-based optimization method.
\item GCN: \cite{GCN}: GCN deploys graph convolution operation to graph data.
\item GAT: \cite{GAT}: GAT leverages the attention mechanism to propagate information of neighbors adaptively. 
\item FAGCN: \cite{FAGCN}: FAGCN is a modified GNN framework that captures signals in different frequencies for representation learning.
\item GPR-GNN: \cite{GPR-GNN}: GPR-GNN integrates generalized PageRank and GNN to assign weight to neighboring nodes and capture representative node features.
\item CARE-GNN: \cite{CARE-GNN}: CARE-GNN utilizes neighboring selection strategies to alleviate fraudulent camouflage for fraud detection problems.
\item $H^{2}$-FDetector: \cite{H2-FDetector}: $H^{2}$-FDetector adopts different propagation rules for homophilic and heterophilic connections in the graph by distinguishing the attention weights of neighbors.
\item BWGNN: \cite{BWGNN}: BWGNN employs beta wavelet transformation to filter the high-frequency signals in fraudsters.
\item SEFraud: \cite{SEFraud}: SEFraud utilizes mask strategies to obtain representative information about fraudsters.
\item CIES-GNN: \cite{ZhangLT24}: CIES-GNN leverages node and structural features to reconstruct a dense subgraph.
\item DOS-GNN: \cite{DOS-GNN}: DOS-GNN integrates oversampling and dual-channel feature aggregation to detect fraudsters.
\item GFAN: \cite{GFAN}: GFAN exploits the semantic information between instances and leverages co-training strategies to enhance camouflaged fraudulent representations.
\end{itemize}
\begin{table*}[!htbp]
	\centering
	\caption{Performance of the Proposed DHMP Model and Comparative Model on Three Datasets. All results are in \%.}
	\label{Performance}
  \resizebox{0.98\textwidth}{!}{
	\begin{tabular}{c c| c c c c | c c c c| c c c c}
		\hline
		\multirow{2}{*}{Method} & Dataset & \multicolumn{4}{c}{YelpChi} & \multicolumn{4}{c}{Amazon}& \multicolumn{4}{c}{FDCompCN}\\
		\cline{2-14}
		~& Metric & Recall & F1-macro & AUC & GMean & Recall & F1-macro & AUC & GMean & Recall & F1-macro & AUC & GMean\\\hline
		\multirow{8}{*}{Baselines} & XGBoost & 19.15 & 61.72 & 59.01 & 43.51 & 69.09 & 72.68 & 79.54 & 78.87 & 61.25 & 61.17 & 50.64 & 58.04\\
		~ & MLP & 69.37 & 61.48 & 77.43 & 70.73 & 78.18 & 72.95 & 87.78 & 82.93& 57.08 & 54.80 & 43.06 & 58.48\\
		\cline{2-14}
            ~ & GCN & 77.53 & 36.67 & 59.33 & 49.46 & 80.00 & 56.43 & 84.61 & 73.72& 52.92 & 51.01 & 40.89 & 43.95\\
            ~ & GAT & 62.15 & 42.77 & 56.13 & 53.13 & 80.00 & 71.46 & 88.03 & 83.04& 52.55 & 51.36 & 38.20 & 42.94\\
		~ & GPRGNN & 75.16 & 57.34 & 77.12 & 69.84 & 80.09 & 64.15 & 89.08 & 82.32& 56.40 & 47.52 & 50.31 & 52.09\\
		~ & FAGCN & 70.64 & 61.11 & 77.90 & 70.86 & 81.21 & 69.30 & 90.48 & 84.33& 57.90 & 48.48 & 51.59 & 49.50\\
		\cline{2-14}
		~ & CARE-GNN & 72.32 & 60.40 & 77.41 & 70.86 & 75.76 & 70.45 & 86.19 & 81.71& 57.21 & 43.59 & 49.00 & 50.10\\
            ~ & $H^{2}$-FDetector & 84.61 & 70.78 & 88.90 & 81.64 & 82.12 & $\underline{71.36}$ & 89.84 & 84.29& 55.96 & 48.33 & 47.89 & 49.10\\
            ~ & BWGNN & 82.56 & 72.32 & 89.72 & 81.92 & 83.94 & 69.43 & 91.91 & 84.67& $\underline{58.41}$ & 47.91 & $\underline{60.79}$ & 55.33\\
            ~ & CIES-GNN & 80.53 & 72.85 & 88.84 & 80.61 & 86.73 & 68.50 & 90.99 & 82.41& 56.99 & 50.58 & 52.53 & 53.32\\
            ~ & DOS-GNN & 82.14 & 70.46 & 81.15 & 81.66 & 85.12 & 69.53 & 91.35 & 83.94 & 57.36 & 49.21 & 52.47 & 52.96\\
            ~ & SEFraud & 78.64 & 72.51 & 86.77 & 82.44 & $\underline{88.67}$ & 71.28 & 90.50 & $\underline{85.13}$ & 57.49 & $\underline{51.31}$ & 50.41 & 53.74\\
            ~ & GFAN & $\underline{82.42}$ & $\underline{73.65}$ & $\underline{90.69}$ & $\underline{83.66}$ & 86.97 & 71.01 & $\underline{91.96}$ & 84.88 & 58.32 & 50.22 & 58.46 & $\underline{55.71}$\\\hline
		Ours & DHMP & $\mathbf{84.39}$ & $\mathbf{74.35}$ & $\mathbf{91.80}$ & $\mathbf{83.68}$ & $\mathbf{88.88}$ & $\mathbf{72.63}$ & $\mathbf{92.32}$ & $\mathbf{86.88}$ & $\mathbf{59.60}$ & $\mathbf{51.63}$ & $\mathbf{61.43}$ & $\mathbf{57.75}$ \\\hline
	\end{tabular}}
\end{table*}

\begin{table*}[!htbp]
	\centering
	\caption{Performance of the ablation experiments on three datasets. All results are in \%.}
	\label{ablation Performance}
  \resizebox{0.98\textwidth}{!}{
	\begin{tabular}{c| c c c c | c c c c| c c c c}
		\hline
		\multirow{2}{*}{Variants} & \multicolumn{4}{c}{YelpChi} & \multicolumn{4}{c}{Amazon}& \multicolumn{4}{c}{FDCompCN}\\
		\cline{2-13}
		~ & Recall & F1-macro & AUC & GMean & Recall & F1-macro & AUC & GMean & Recall & F1-macro & AUC & GMean\\\hline
		DHMP$_{sep}$ & 76.32 & 68.58 & 84.64 & 74.36 & 81.20& 64.31& 82.51& 73.53& 50.01& 44.31& 51.52& 48.79\\
		DHMP$_{homo}$ & 78.31 & 67.78 & 83.42 & 76.73 & 80.07& 64.71& 84.90& 72.24& $\underline{52.94}$ & 44.84 & 48.35 & 47.62\\
            DHMP$_{heter}$ & 78.53 & 66.65 & 82.36 & 79.88 & 82.44 & 62.18& 81.14& 75.53& 49.34& 42.81& 50.23& 48.78\\
		DHMP$_{rel}$ & $\underline{80.64}$ & $\underline{69.15}$ & $\underline{87.98}$ & $\underline{74.86}$ & $\underline{83.64}$ & $\underline{69.54}$& $\underline{87.54}$& $\underline{81.21}$& 52.86 & $\underline{48.83}$& $\underline{56.61}$& $\underline{53.93}$\\\hline
		DHMP & $\mathbf{84.39}$ & $\mathbf{74.35}$ & $\mathbf{91.80}$ & $\mathbf{84.38}$ & $\mathbf{88.88}$ & $\mathbf{72.63}$ & $\mathbf{92.32}$ & $\mathbf{86.88}$ & $\mathbf{59.60}$ & $\mathbf{51.63}$ & $\mathbf{61.43}$ & $\mathbf{57.75}$ \\\hline
	\end{tabular}}
\end{table*}

\begin{figure}[!htb]
\centering
\resizebox{0.48\textwidth}{!}{
\subfloat{
	\includegraphics[scale=0.2]{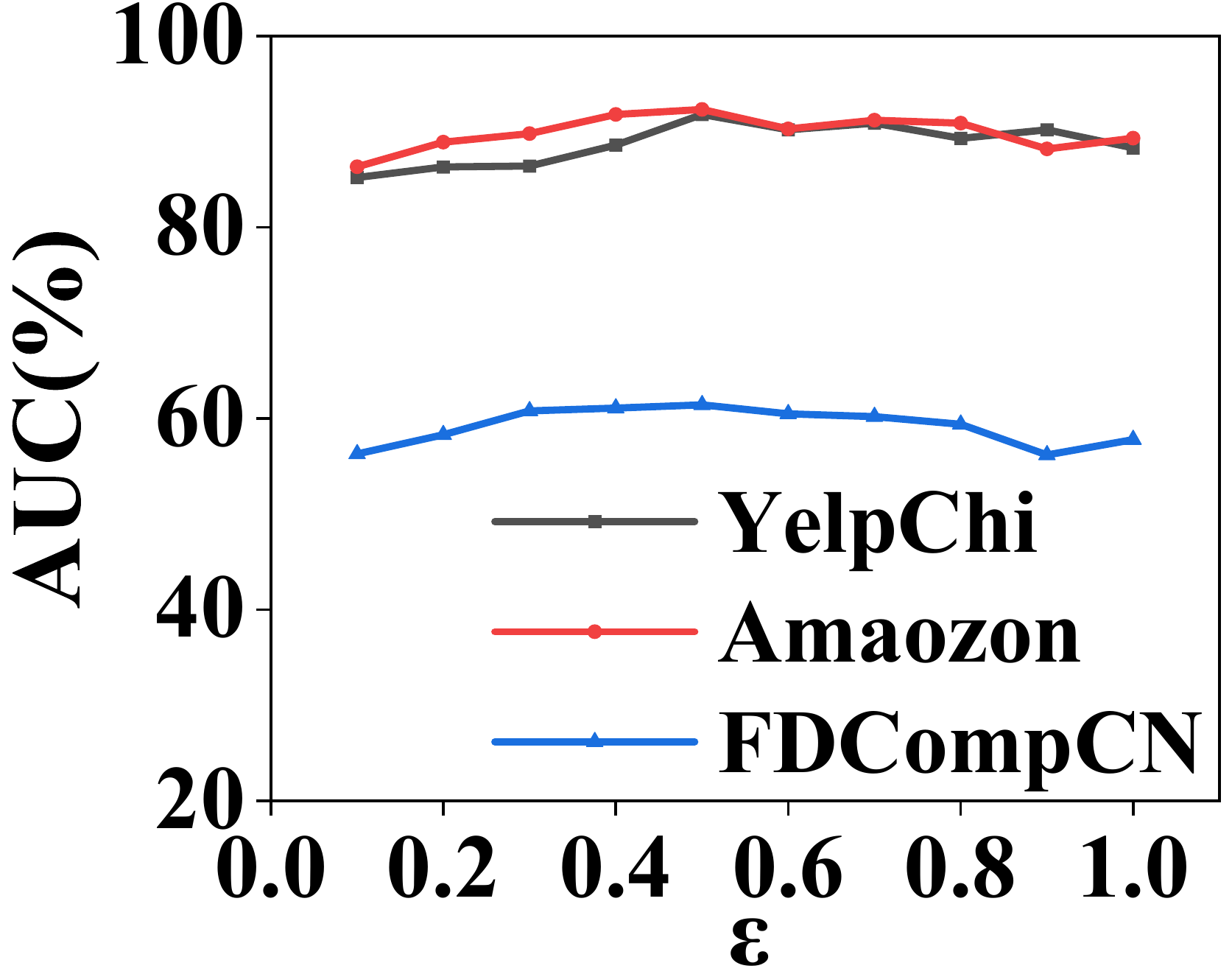} \label{epsilon AUC}}
\subfloat{
		\includegraphics[scale=0.2]{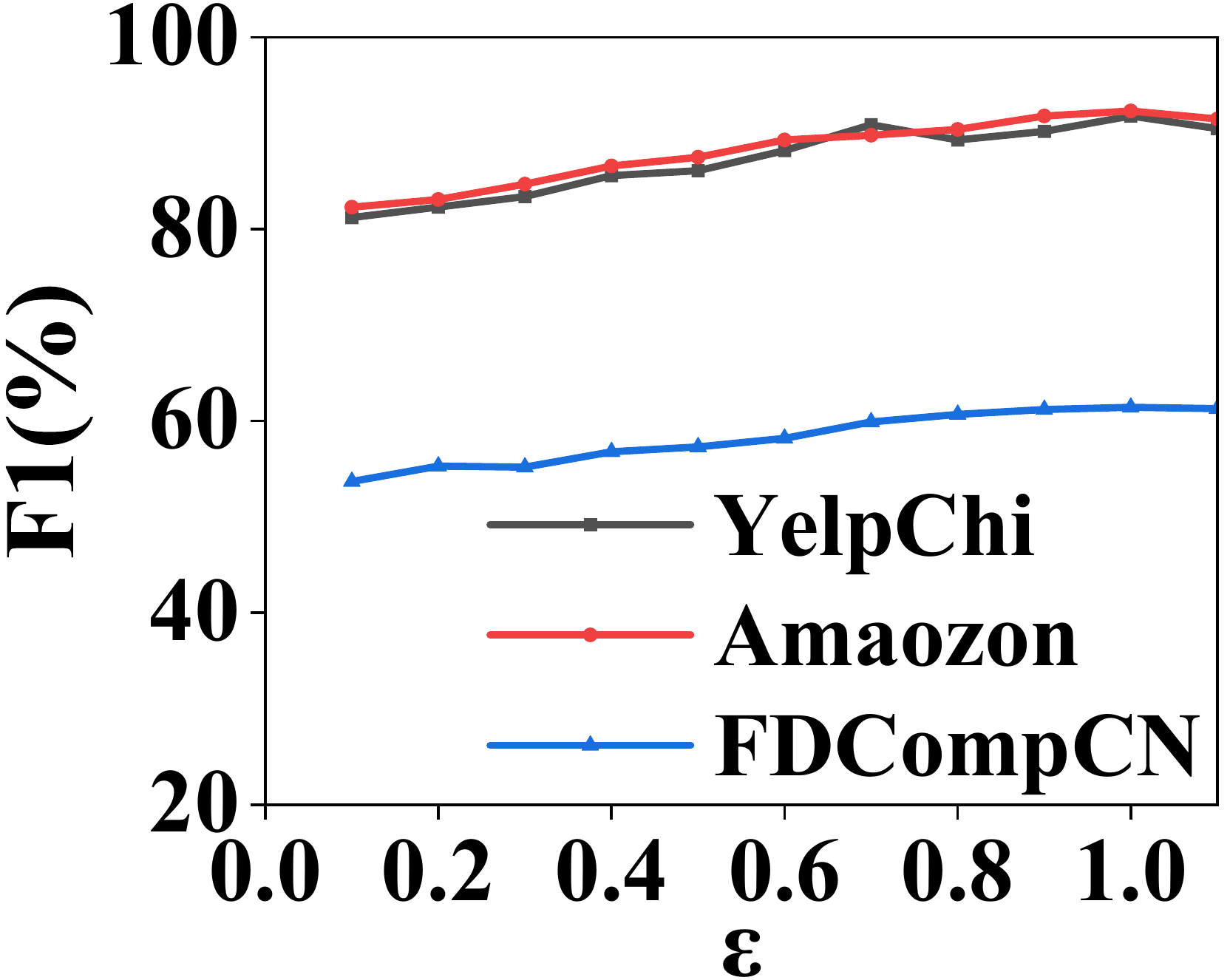} \label{epsilon F}}}
\caption{The results of sensitivity experiments of hyperparameter $\epsilon$: (a) AUC of DHMP on three datasets;  (b) F1-macro of DHMP on three datasets}
\label{epsilonsensitivity}
\end{figure}
\begin{figure}[!htb]
\centering
\resizebox{0.48\textwidth}{!}{
\subfloat{
	\includegraphics[scale=0.2]{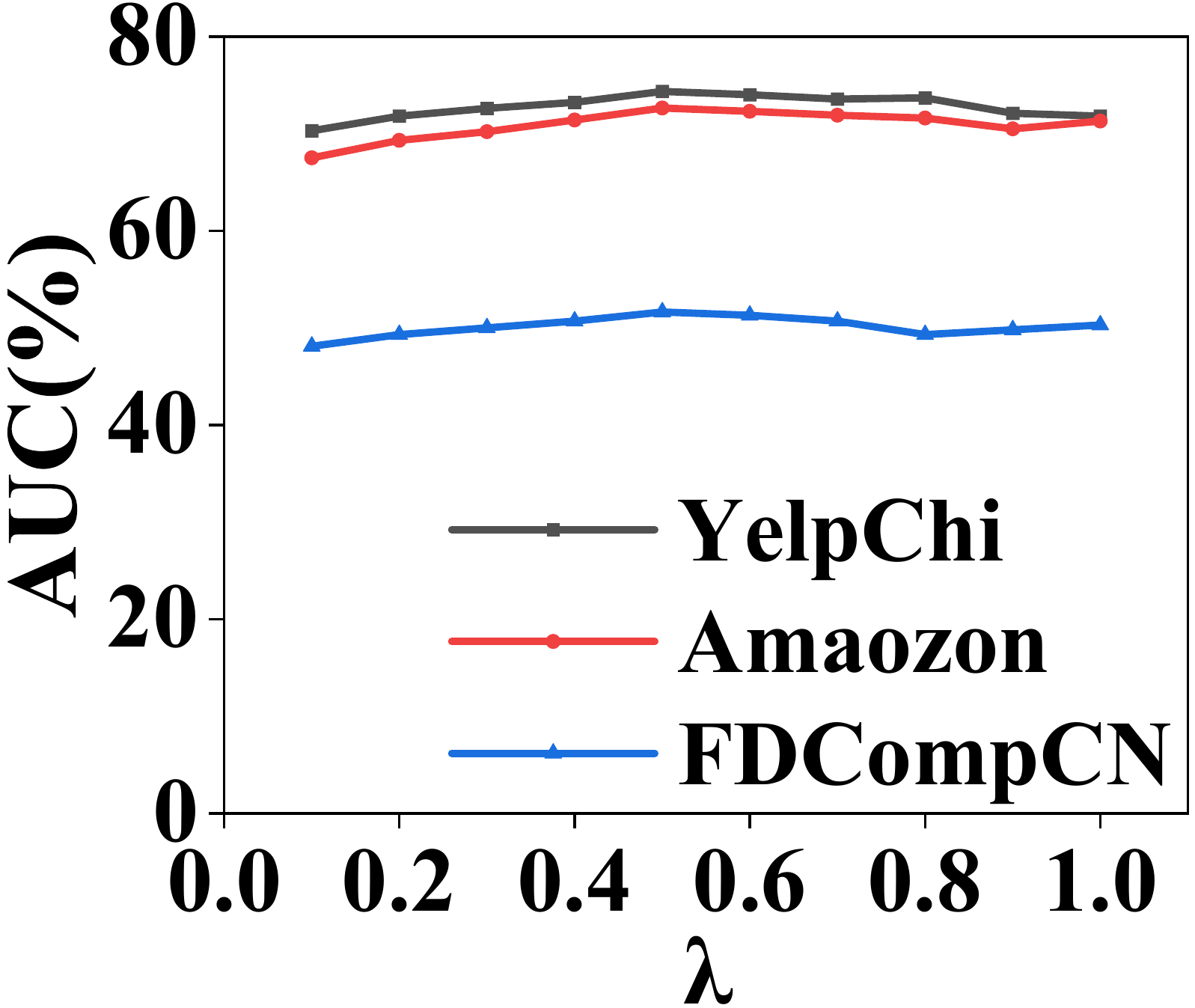} \label{lambda AUC}}
\subfloat{
		\includegraphics[scale=0.2]{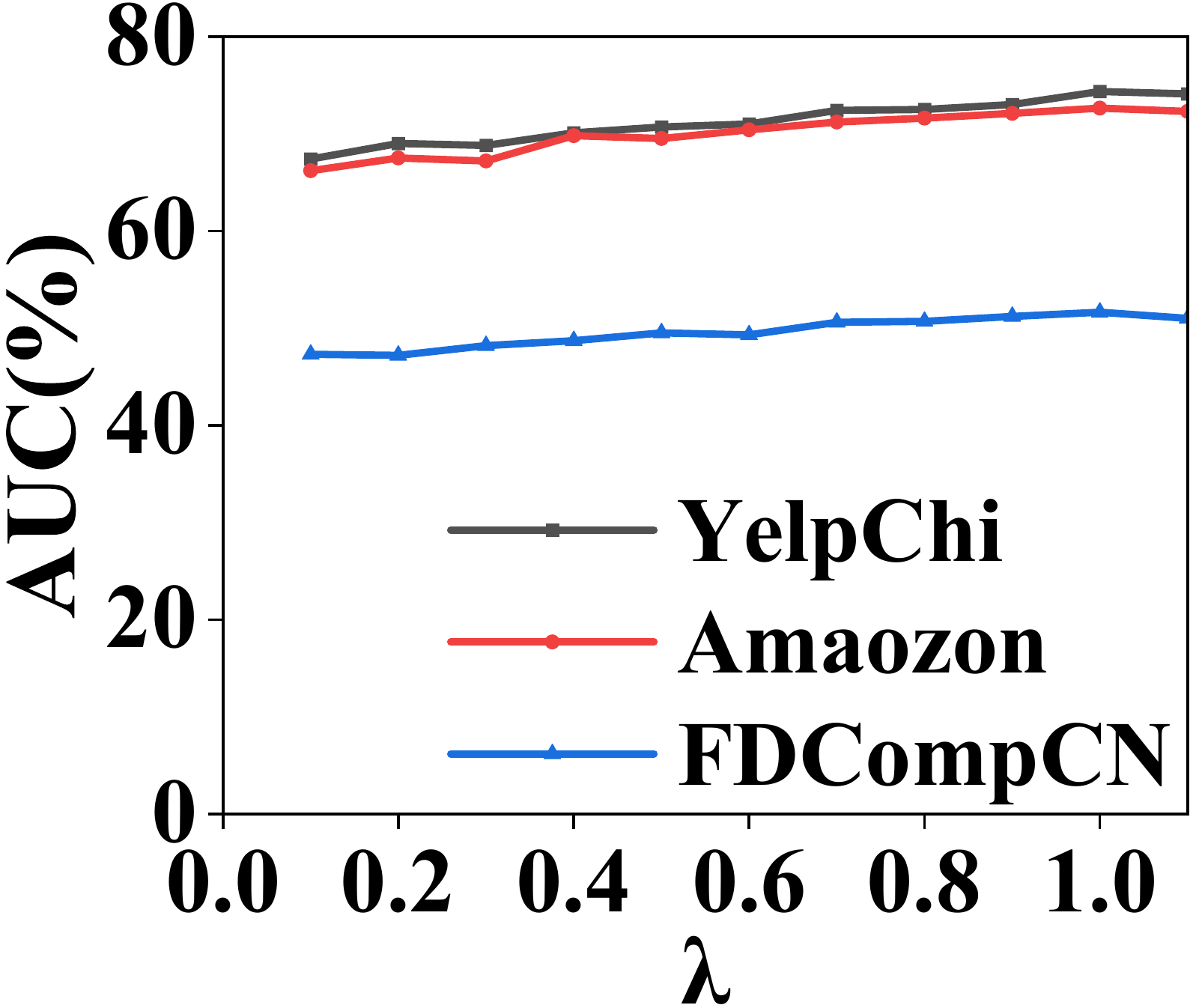} \label{lambda F}}}
\caption{The results of sensitivity experiments of hyperparameter $\lambda$: (a) AUC of DHMP on three datasets;  (b) F1-macro of DHMP on three datasets}
\label{lambdasensitivity}
\end{figure}
\subsubsection{Evaluation settings}
Because of the obvious data imbalance in fraud detection, we choose four metrics to assess all models: AUC, Recall, GMean, and F1-score. AUC describes the prediction results by probabilities, Recall reflects the model's ability to capture anomalies, GMean comprehensively evaluates the balance between Recall and accuracy, and F1-score integrates Recall and Precision.

\subsubsection{Implementation details}
The computer configuration is an NVIDIA A40 GPU, 40GB of RAM, and a 2.60GHz Xeon (R) Gold 6240 CPU. The experiments are conducted with PyTorch in Python 3.9.12, and all the baselines can be reproduced using public codes.

DHMP is trained for 3000 epochs with a patience of 200. The hyperparameter $\lambda$ and $\epsilon$ are set to 1 and 0.5, respectively. Adam optimizer is used to train the model, respectively, with the learning rate 0.01 for the YelpChi dataset, 0.1 for the Amazon dataset, and 0.001 for the FDCompCN dataset. The weight decay is set to 0.00005. The dimension of node representation is 8. The dropout rate is set to 0.1. We leverage the best parameters for all baselines according to the corresponding authors.
\begin{figure*}[!htb]
\centering
\subfloat[GCN]{
	\includegraphics[scale=0.26]{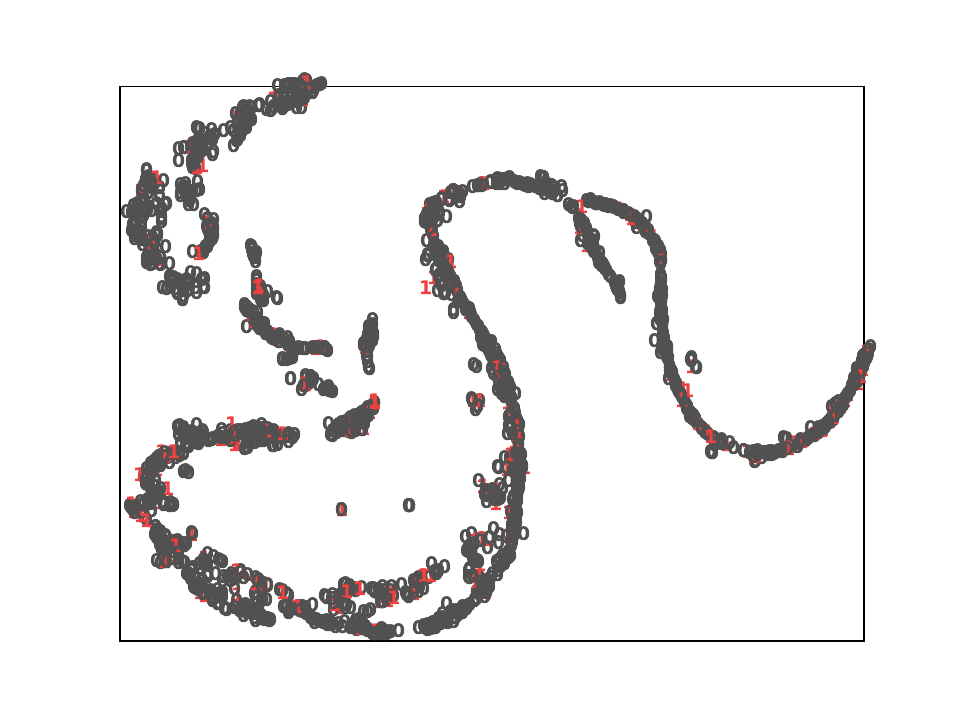} \label{figGCN}}
\subfloat[$H^{2}$-FDetector]{
		\includegraphics[scale=0.26]{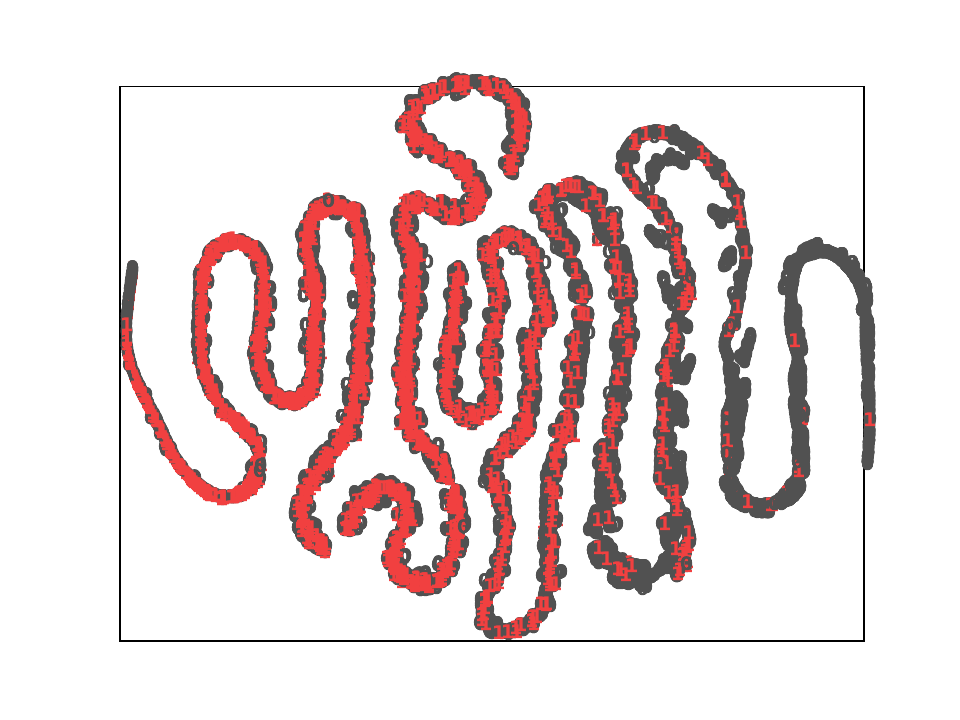} \label{figH2}}
\subfloat[BWGNN]{
		\includegraphics[scale=0.26]{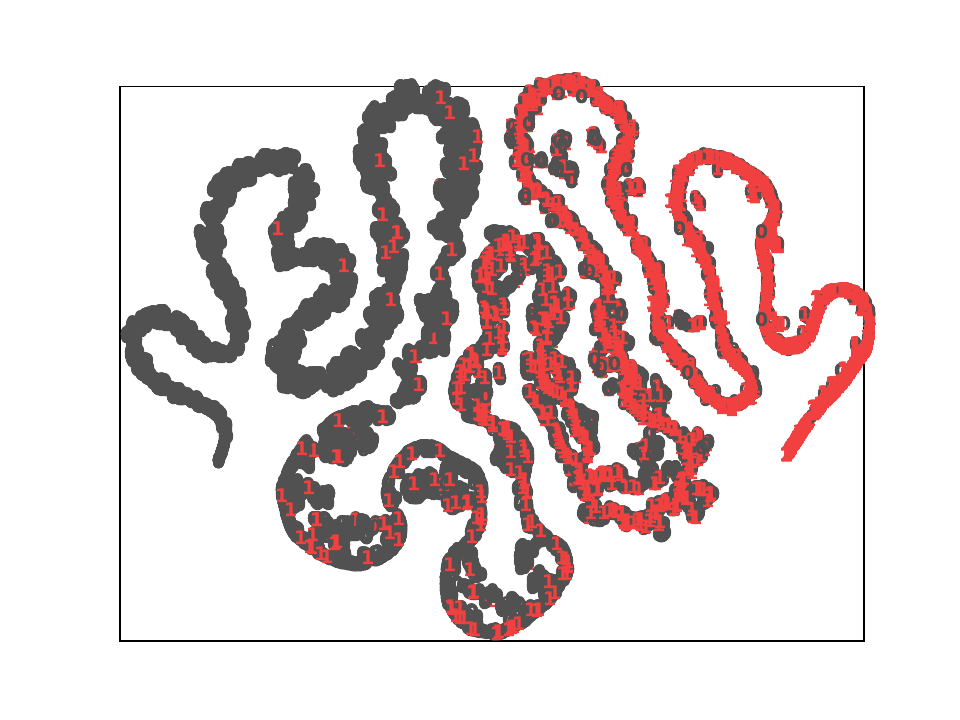} \label{figBW}}
\subfloat[GFAN]{
		\includegraphics[scale=0.26]{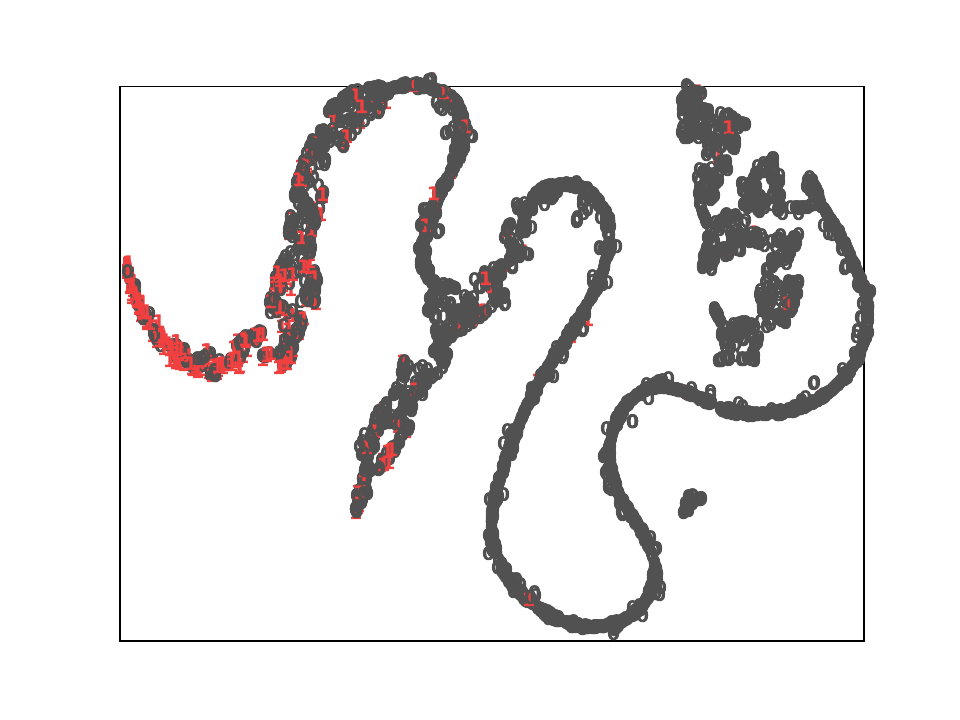} \label{figGFAN}}
\subfloat[DHMP]{
		\includegraphics[scale=0.26]{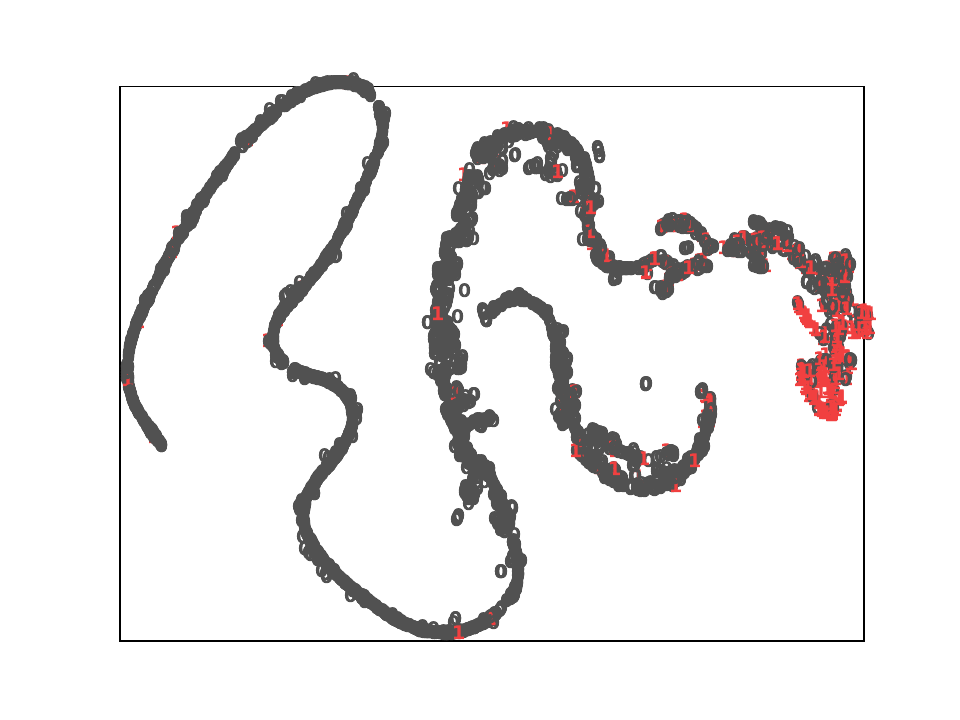} \label{figDHMP}}
\caption{The results of visualization of the representation learning on the YelpChi dataset.}
\label{Visualization}
\end{figure*}

\subsection{Overall Performance}
In this section, we answer \textbf{RQ1} to demonstrate the effectiveness of DHMP. The performance observations are shown in Table \ref{Performance}. From Table \ref{Performance}, shallow methods display poorer performance than deep learning methods because they fail to capture latent representation space. In contrast, GNNs have relatively remarkable observations because they can exploit complex feature representations through nonlinear transformation and topological interaction. Furthermore, GNN-based fraud detection approaches exhibit superiority over traditional GNNs. This can be attributed to the ability to resist data imbalance problems in fraud detection and more effectively extract characteristics of minor categories. Even so, the proposed DHMP achieves state-of-the-art performance over all baselines. Take the YelpChi dataset as an example. Compared to shallow methods, DHMP improves Recall, F1-macro, AUC, and GMean by at least 15.02\%, 12.63\%, 14.37\%, and 12.95\%. Compared to GNNs, DHMP improves Recall, F1-macro, AUC, and GMean by at least 6.86\%, 13.24\%, 13.90\%, and 12.82\%. Compared to GNNs-based fraud detection approaches, DHMP improves Recall, F1-macro, AUC, and GMean by at least 1.97\%, 0.7\%, 1.11\%, and 0.02\%. Similarly, we can see an obvious improvement in DHMP over baselines on the other two public datasets.

In summary, with the core advanced idea that independently exploits heterophilic and homophilic latent features, DHMP can capture valuable fraudulent characteristics and accurately identify fraudulent individuals. Specifically, the advancement of performance illustrates the effectiveness of the heterophily separation module in DHMP, which can perceive and split heterophilic connections in the original graph. Furthermore, it also confirms that the re-scaled residual message aggregation can extract distinct latent features and boost model discrimination capacity. Last, the multi-scale information fusion, including heterophilic and homophilic information fusion and relational information fusion, can help the model derive comprehensive embeddings of nodes, which enables the training of a robust and stable fraud detection framework to identify fraudsters with camouflage behavior effectively.

\subsection{Ablation experiments}
In this section, we would like to respond \textbf{RQ2} to verify the utility of each component in the DHMP through ablation experiments. We investigate the effectiveness of components in DHMP and generate four variants: DHMP$_{sep}$, DHMP$_{homo}$, DHMP$_{heter}$, and DHMP$_{rel}$. 

DHMP$_{sep}$ represents DHMP without heterophily separation module, and without heterophily separation, we only use one channel for re-scale residual message aggregation. DHMP$_{homo}$ and DHMP$_{heter}$ respectively denote DHMP without a homophilic propagation module and DHMP without a heterophilic propagation module. For these two variants, the subgraph information fusion module is excluded because the variants don't capture signals with different frequencies. DHMP$_{rel}$ DHMP without relation representation fusion module, which means the variant is trained on the homogeneous graph.

From the observations of ablation experiments in Table \ref{ablation Performance}, all compositions contribute to the comportment of DHMP. Specifically, the heterophily separation module, homophilic propagation module, and heterophilic propagation module have relatively larger contributions to the performance, demonstrating the significance of independently extracting high-pass and low-pass signals in the graph. In addition, relation discrimination also helps to capture key information about the fraudsters because fraudsters behave differently under different relations. This result is under our anticipations. The fundamental operational capability of DHMP hinges on the effective extraction of signals with different frequencies and their synergistic integration, thereby mitigating the adverse effects of fraudulent camouflage within relational data.

\subsection{Sensitivity experiments}
In response to \textbf{RQ3}, we further conduct sensitivity experiments of DHMP. First, we execute sensitivity experiments of hyperparameter $\epsilon$, and the results are shown in Fig. \ref{epsilonsensitivity}. We can conclude that too large and small values of $\epsilon$ constrain the performance. Smaller $\epsilon$ suggests less preservation of the original feature of the neighbors during message passing, resulting in the deficiency in maintaining the long-term dependence of initial input. As a result, performance declined slightly. In contrast, a value that is too large may impact the ability of the model to learn latent representation spaces, which also results in limited discriminative ability. According to the observation, the best value is 0.5.

Then, we study the influence of $\lambda$ on the performance, which can be observed in Fig. \ref{lambdasensitivity}. $\lambda$ takes control of the perception of heterophily in the graph. According to the reports, the performance of AUC and F1-macro improves as $\lambda$ increases because a large value indicates that the model has great power to partition signals with different frequencies. Nevertheless, too large a value results in a relative increase in loss, which 
also have a light impact on the performance.

\subsection{Visualization}
To answer \textbf{RQ4}, we substantiate that DHMP yields more discernible node embeddings by conducting an exhaustive visualization analysis, underscoring the distinctions between the embeddings produced by DHMP and four selected state-of-the-art benchmark approaches. Given the constraints imposed by page limitations, we have chosen to focus our presentation on the YelpChi dataset, thereby showcasing the comparative performance in a representative and illustrative manner. The reports are shown in Fig. \ref{Visualization}, where the red marks indicate benign instances, and the blue marks denote fraudulent embeddings.

According to the observations, GCN fails to generate clear clusters between fraud and non-fraud points and suffers from fuzzy boundaries. In contrast, $H^2$-Detector and BWGNN can roughly separate features of different categories. DHMP exhibits better-distinguished boundaries with higher cohesion. GFAN and DHMP form clear boundaries and redistribute the nodes, making them further apart. However, GFAN learns an irregular and limited latent vector space, which has a disadvantageous impact on the model's generalization performance. It is worth noting that DMPH can significantly compress the feature space dimension of fraudulent classes and effectively condense their essential features. Moreover, it demonstrates a remarkable generalization capability for a broad spectrum of benign users.

\section{Conclusion}
In this paper, we develop DHMP for fraud detection. DHMP first leverages a heterophily separation module to split the original graph and then utilizes a re-scaled residual message aggregation module to capture signals in different frequencies. Last, DHMP integrates multi-relation embeddings for fraudster prediction. Extensive experiments verify the superiority of DHMP over state-of-the-art baselines. In the future, we will continue to develop more effective methods for identifying complex fraudulent behaviors and further improve the scalability of the models.
\section*{Acknowledgment}
This work was supported by the National Natural Science Foundation of China under Grant 72210107001, the Beijing Natural Science Foundation under Grant IS23128, the Fundamental Research Funds for the Central Universities, and by the CAS PIFI International Outstanding Team Project (2024PG0013).
\bibliographystyle{IEEEtran}
\bibliography{ref}
\end{document}